\documentclass[letterpaper, 10 pt, conference]{ieeeconf}  

\IEEEoverridecommandlockouts                              

\usepackage{amsmath}
\usepackage{graphicx}
\usepackage[colorlinks=true, allcolors=blue]{hyperref}
\usepackage{tabularx}
\usepackage{subcaption}

\title{\LARGE \bf
Bootstrapping Self-Supervised Learning of Binary Classification Using Error Bounds: A Case Study on a Robotic Insertion Task
}

\author{Zebin Duan$^{1}$, Norbert Krüger$^{1, 2}$, Juan Heredia$^{1}$, Thorbjørn Mosekjær Iversen$^{1}$, Frederik Hagelskjær$^{1}$ 
    \thanks{
    $^{1}$All authors are with the Mærsk Mc-Kinney Møller Institute, University of Southern Denmark, 5230 Odense M, Denmark.
            {\tt\small \{zeb,norbert,jehm,thmi,frhag\}@mmmi.sdu.dk}
    }%
    \thanks{$^{2}$The author is affiliated with the Danish Institute for Advanced Study (DIAS), University of Southern Denmark, 5230 Odense M, Denmark. {\tt\small norbert@mmmi.sdu.dk}}%
}

\begin{document}

\maketitle
\thispagestyle{empty}
\pagestyle{empty}

\begin{abstract}
Flexible manufacturing requires rapid deployment of solutions and minimal setup time to remain competitive. An essential attribute is the ability to control error levels, as failures can range from minor performance degradation to severe equipment damage. However, conventional deployment often involves extensive setup, data collection, model training or parameter tuning, and system testing, resulting in significant delays that hinder commercial feasibility.
We propose a data engine which gathers data and improves its performance while executing the task. 
The data engine consists of two classifiers, a fast model prediction and expensive verification.
First, a model prediction is performed and based on the confidence level of the prediction, the expensive verification can be used. By adjusting the confidence level, users can control the level of tolerable error.

Our method is implemented on a real-world robotic insertion task, which uses force data for the model prediction. The system applies UMAP dimensionality reduction and uses Wilson-Score to compute the confidence bounds of the prediction. 
Results demonstrate the ability to learn and reduce the need for expensive verifications over time, while staying within the set error-rate. 
The results highlight the potential of confidence bounds in self-improving models to enhance reliability in robotic classification task.
\end{abstract}

\section{Introduction}


Robotic systems are key drivers in flexible manufacturing solutions, since they can be configured for new tasks without large mechanical changes. Robotic solutions are generally programmed as set of consecutive operations. As each operation is dependent on the success of the previous action, each action is expected to complete without error. Otherwise, such errors will result in failures in the subsequent operations, and thus a failure of the overall task. Depending on the setup, the severity a failure can vary from increased run-time to damaging the setup. 

\begin{figure}[t]
    \centering
    \includegraphics[width=0.99\linewidth]{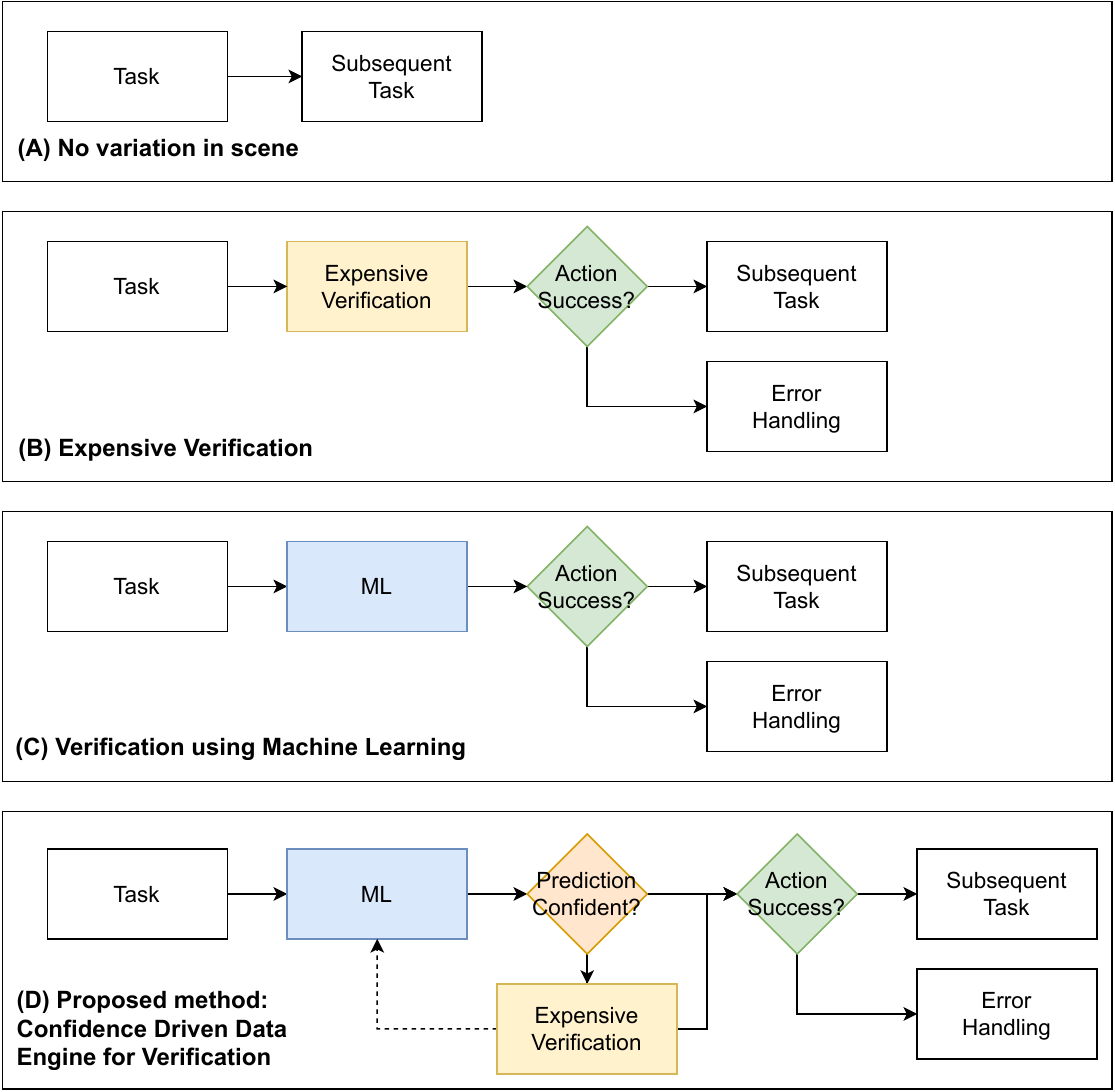}
    \caption{The different types of verification systems for robotic actions. From the top: Task without any variation, and thus no verification is necessary. Using an expensive verification method to ensure success. The system then also needs different paths to handle failure cases. Then using a machine learning model for verification instead of the expensive solution. Finally, at the bottom, our proposal for a data engine driven by the confidence of the prediction. First the machine learning model makes a prediction, if the confidence of the prediction is adequate we rely on the machine learning model, otherwise the expensive verification is used. The result of the expensive verification is then collected and used for training the machine learning model.
    }
    \label{fig:enter-label}
\end{figure}

To ensure reliability, robot systems are often a sequence of simple operations without any kind of adaptation, see Fig.~\ref{fig:enter-label} (A). This requires large amounts of manual work, fine tuning and long set-up times, making systems commercially unfeasible.

Alternatively, errors can be detected with a verification step, which could be human intervention \cite{boschetti2023improving} or by the robot system itself \cite{hagelskjaer2025good}. While ensuring robustness, this verification step is often expensive in run-time making the system less profitable. We call this verification step the {\it expensive verification}. This approach is shown in Fig.~\ref{fig:enter-label} (B). 

\begin{figure*}[t]
    \vspace{3mm}
    \centering
    \includegraphics[width=0.95\linewidth]{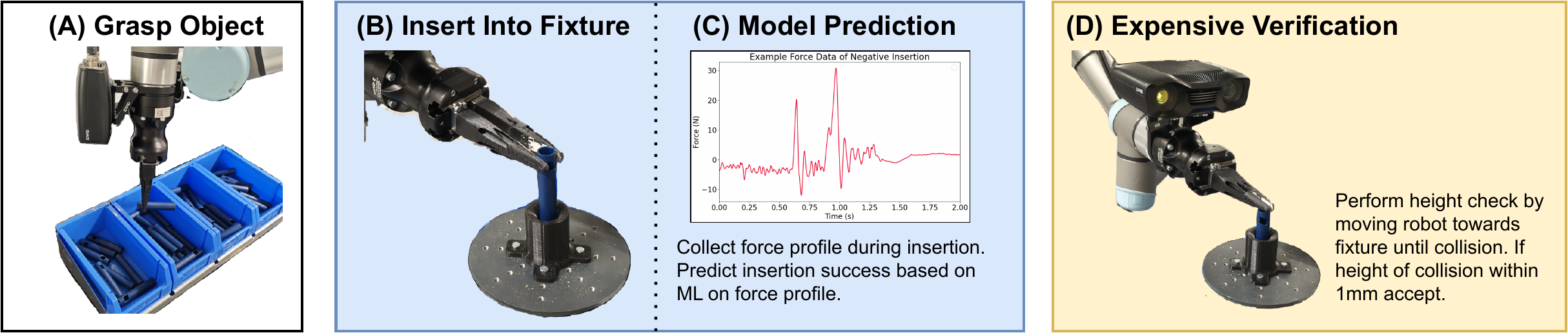}
    \caption{ The two different verification methods implemented in our system. (A) First the object is grasped from a bin. (B) It is then placed into the fixture and using either the (C) model prediction or (D) the expensive verification to classify success of the action. If the action is a success, the object is moved further into the production line, otherwise it is moved back to the bin and a new grasp and insertion is performed.}
    \label{fig:realsystem}
\end{figure*}

As the performance of machine learning models has grown in recent years, so has their usage in verification \cite{twala2009robot, alonso2016bayesian, nentwich2020data, elia2025anomaly}. By using a model prediction, it is possible to avoid the expensive verification. This approach is illustrated in Fig.~\ref{fig:enter-label} (C). 
%
Examples of both expensive verification and model prediction are applied in the system used in this paper. 
Our system (see Fig. \ref{fig:realsystem}) is based on the configuration presented in \cite{hagelskjaer2025good, duan2025towards}. 
A component is grasped from a bin (see Fig. \ref{fig:realsystem} (A)) and moved to a fixture (see Fig. \ref{fig:realsystem} (B)) for further production. The verification step is introduced to ensure that the object is placed correctly in the fixture. The expensive verification (see Fig. \ref{fig:realsystem} (D)) is a touch up that ensures the object is placed in the correct height. This verification is simple to program and reliable, however, it is slow to execute and adds significantly to the run-time. 

As an alternative, a machine learning model is trained on the force profile obtained during the insertion. While much faster at run-time, the machine learning model requires large amounts of data sampling and finetuning until it is reliable enough to perform. 
This limits robot systems to either very simple operations without any variation in the scene, tasks where run-time is not critical to profitability, or tasks where very large production batches allow for extended setup and test time.

An alternative strategy is to use a data engine to continuously collect data and improve performance of models. Here the expensive verification can be used initially, and as data is collected, the machine learning model can be gradually introduced. 
However, while the model can be trained on data collected from the expensive verification it is not easy to determine when the model has gathered enough data to perform reliably enough when no guarantees for the frequency of errors can be provided. 

In this paper, we introduce a data engine based on confidence scores from the model prediction. This method is depicted in Fig.~\ref{fig:enter-label} (D). Using Wilson-Score \cite{wilson1927probable} for classification has shown to successfully provide guaranteed bounds for error frequency \cite{iversen2025wskde}. In this work, we integrate the Wilson-Score into a data engine to provide not only a classification for the machine learning model, but also bounds on the probability of the prediction error. 
If the confidence of the model is below a set threshold, we rely on the expensive verification and the data generated is fed to the model. As data is collected, the prediction quality of the model increases and the expensive verification is used less frequently. If the model encounters data not fitting with the current model, the expensive verification is used instead of making an uncertain prediction. 

We demonstrate that the confidence level from the Wilson-Score method can be used to adjust the error frequency. The system is able to run immediately, relying on the expensive verification until the confidence of the model is large enough that the error frequency is not violated.





% ...


The main contributions presented in this paper are:

\begin{itemize}
    \item We establish online-learning in a working robot system where upper error bounds are guaranteed by means of the Wilson-Score.
    \item We investigate different parameters settings of the error bound and show their effect for learning.
    \item We compare our approach with a baseline based on the Binomial Interval Score.
\end{itemize}







\section{Related Work}


First we give an overview of methods for prediction of failures in systems similar to ours. Then methods for calculating confidence in binary classification are reviewed and as we use dimensionality reduction to pre-process the data, we report on related work in these areas.

\subsection{Failure prediction in assembly processes}

As disruptions in manufacturing are very costly, detection of failures has been studied extensively \cite{elia2025anomaly}. One approach to avoid faults is to predict anomalies using sensor data \cite{twala2009robot, alonso2016bayesian, nentwich2020data}. 

A method similar to ours is presented in \cite{alonso2016bayesian}, where also force data is recorded during an insertion operation and number of  classifiers for error prediction are tested. They show good classification results on the UCI benchmark dataset \cite{robot_execution_failures_138} and a Naive Bayesian classifier is found to work best. However, they don't use any kind if confidence bounds for the classifiers.

Decision trees have also been used for the classification of errors using force data \cite{twala2009robot}.
Similar to \cite{alonso2016bayesian}, the method is tested on the UCI benchmark dataset. We do not use the UCI benchmark dataset as the sampling rate and dataset size are much smaller than in our approach.




These methods all rely on supervised learning of collected data. Thus an experimental set-up separate from the running assembly system is required to collect the training data which is a costly procedure. 

Another solution is to collect data when the system fails \cite{boschetti2023improving}. But, in many cases this is very costly as it will lead to many system failures. However, this can be added to the system as an additional feature. Thus even though, faults are to be avoided, if they occur the recorded data can be used to improve the system. We add this as an additional feature to our system, were any system failures are logged and then labeled to improve the system.

Among existing approaches, \cite{duan2025towards} is most related to this work, using a $k$-NN on force data to classify the success of an insertion task. All $k$ nearest neighbors are required to vote for the same prediction, otherwise the expensive verification is performed. While the system is similar to our method, it uses a simple $k$-NN for the classification. The number percentage of $k$-NN inliers does not translate to a confidence score. The system, therefore, is not able to provide an expected failure rate.

By using the Wilson-Score method for classification our system is able to provide a bound on the maximum frequency of failures allowing for a more reliable process.   




\begin{figure}
    \vspace{3mm}
    \centering
    \includegraphics[width=0.99\linewidth]{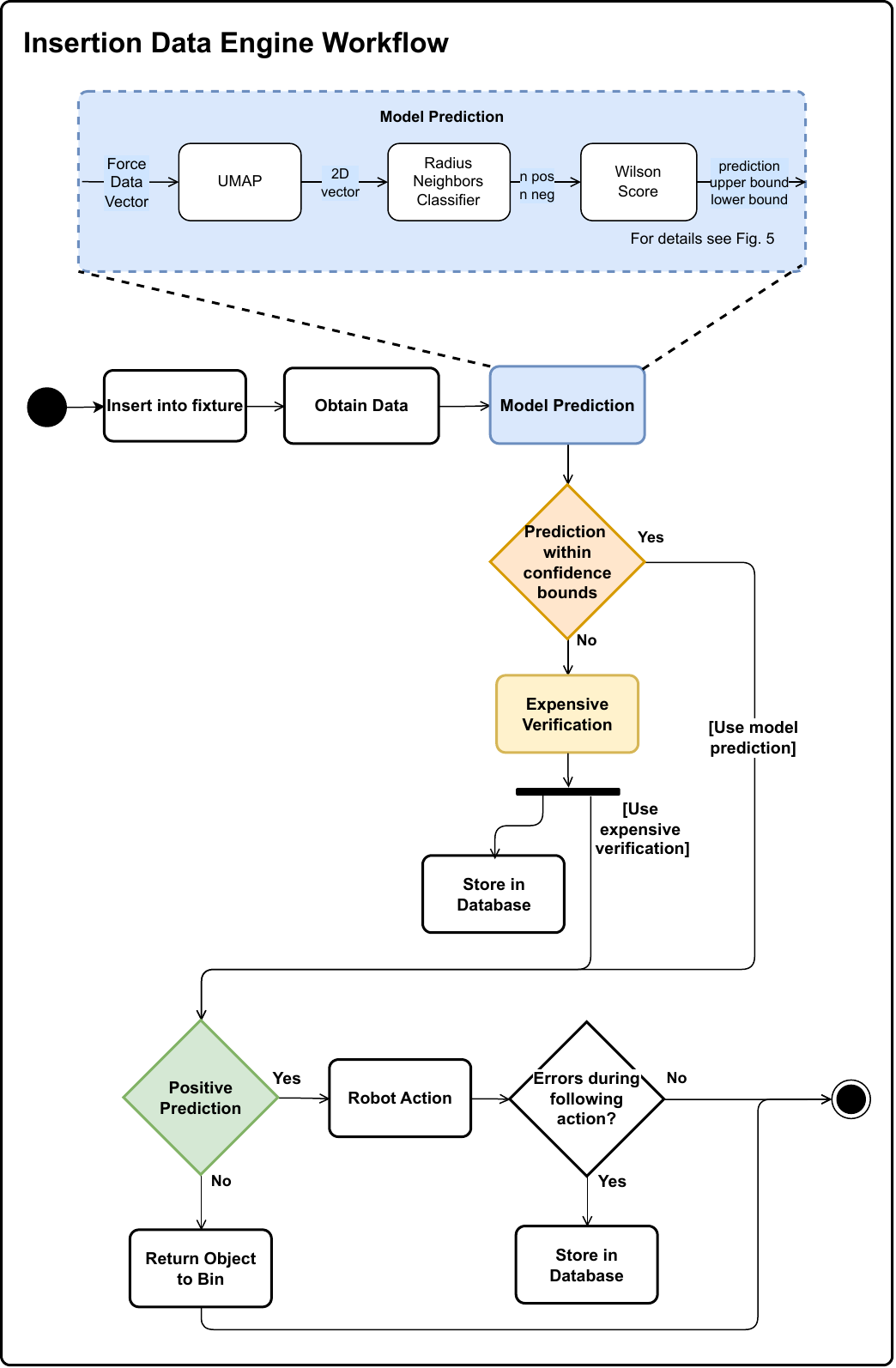}
    \caption{Action diagram of the data engine for an insertion.}
    \label{fig:method}
\end{figure}

\subsection{Confidence Bounds for Binary Classification}
There exists several different methods for quantifying the uncertainty in binary classification. As mentioned the $k$-NN classifier provides a simple but also inaccurate way to estimate the confidence, since it does not take the sampling uncertainty of the neighboring points into consideration.

A more principled method for uncertainty estimation is Gaussian Process Classification~\cite{rasmussen2006gaussian}. While this methodology performs well, it is a computationally expensive method, so approximate methods are required for large data sets, such as those that will be encountered in a running system.

In this work we use a variant of the Wilson Score Kernel Density Classifier \cite{iversen2025wskde}, which is a kernel based method that performs similar to Gaussian process classification, but at a lower computation complexity. Our implementation aggregates the binary outcome of neighboring points using a $k$-NN radius search, and computes the bounds using the Wilson Score method, which makes it equivalent to a Wilson Score Kernel Density Classifier with a uniform kernel instead of a Gaussian. Unlike~\cite{iversen2025wskde}, we choose a fixed radius, as a kernel optimized on a low number of data points, as encountered when the system is starting up, is prone to over-smoothing, which would result in inaccurate confidence bounds.

\subsection{Dimensionality Reduction}
Dimensionality reduction techniques offers several advantages when dealing with high dimensional data, including the removal of insignificant or redundant information, reduced training time, improved model performance, mitigation of overfitting, easier data visualization, and enhanced prediction accuracy, as demonstrated by \cite{sabique2023}, where the authors explored how these techniques could affect force estimation in robotic assisted surgery using neural networks.

They can be used for varying classification tasks when involving time-series data, such as for depression detection \cite{misgar2022}, human hand movement \cite{rabin2020} or sleep stage classification \cite{deng2024}. The authors of \cite{vlaic2025} provide a comparative study of dimensionality reduction techniques for time-series data, analyzing their characteristics and limitations. 


Recent and popular dimensionality reduction techniques include Uniform Manifold Approximation and Projection (UMAP) \cite{2018arXivUMAP}, a manifold learning technique which preserves local neighborhood relationships while also maintaining, to some extent, global structure, and has been shown to outperform several other nonlinear dimensionality reduction methods in these aspects \cite{huang2022}.
For our work, we adopt UMAP, as it is particularly effective for visualizing and interpreting large, high-dimensional datasets.

\section{Method}


The core of our method is an online data engine, which is able to use both a machine learning model and an expensive verification method, for an insertion quality classification use case. This allows the system to minimize incorrect classifications while maintaining efficiency. 

The following section depicts in detail the individual components of the system in Fig. \ref{fig:method}. We begin with the data collection process ("Insert into the Fixture" and "Obtain Data"), followed by the dimensionality reduction technique ("UMAP"). We then perform the adopted machine learning model ("Radius Neighbors Classification") and apply the "Wilson Score" method for computing confidence bounds. 

Finally, we show how these components are integrated into the complete data engine (below "Model Prediction"): when the machine learning model is still too uncertain to perform a prediction (i.e., the prediction is not within the required confidence bounds), the "Expensive Verification" is relied upon.

The purpose of the expensive verification step is to ensure the correct labeling of the robotic operation. In this work, it is realized as a single operation named height check (see Fig. \ref{fig:realsystem} (D)) and described in more details in Sec. \ref{subsec:datacollection}.

Each time the expensive verification is used, the system collects training data to further improve the machine learning model ("Store in Database"). Depending on the result of the prediction, the system executes accordingly and generates further data for improving the model. 
The confidence bounds of the system guarantee a limit on the failure rate while reducing the number of required expensive verifications.

In the sections below, we now describe these steps in detail.

\subsection{Data collection}
\label{subsec:datacollection}

The data was collected from a random bin picking task followed by an insertion task as shown in Fig. \ref{fig:realsystem}. After the insertion, the system needs to verify if the action was successful or not. This can be done using the height check (i.e., the expensive verification in Fig. \ref{fig:realsystem} (D)), and it allows to classify with 100\% certainty between correct and incorrect insertions, at the cost of increased overall time (in our case 5 seconds per cycle). Correct insertions have a fixed height from the fixture, whereas all the other cases (failed, partial or missed insertions) are classified as incorrect.

In general, the expensive verification can be a series of operations, but they need to provide with absolute certainty the perfect label.

During the insertion task, force data has been recorded using the force-torque sensor integrated in the robot end-effector. Each force measurement spans a $2$ second interval sampled at $500$Hz. resulting in a feature vector of $1000$ elements per sample. In this case study, the movement of the robot was mainly along the z-axis. For this reason, only the $z$-axis force data was retained, as incorporating the $x$- and $y$-axis forces not only increased computational time due to input force data being three times it's initial size, but also did not improve performance, indicating that the $z$-axis contains the most discriminative information for the classification task.

\subsection{Dimensionality Reduction}

We first reduce the high-dimensional feature space using UMAP to obtain a two-dimensional representation.

\begin{figure}[thb]
    \centering
    \includegraphics[width=0.99\linewidth]{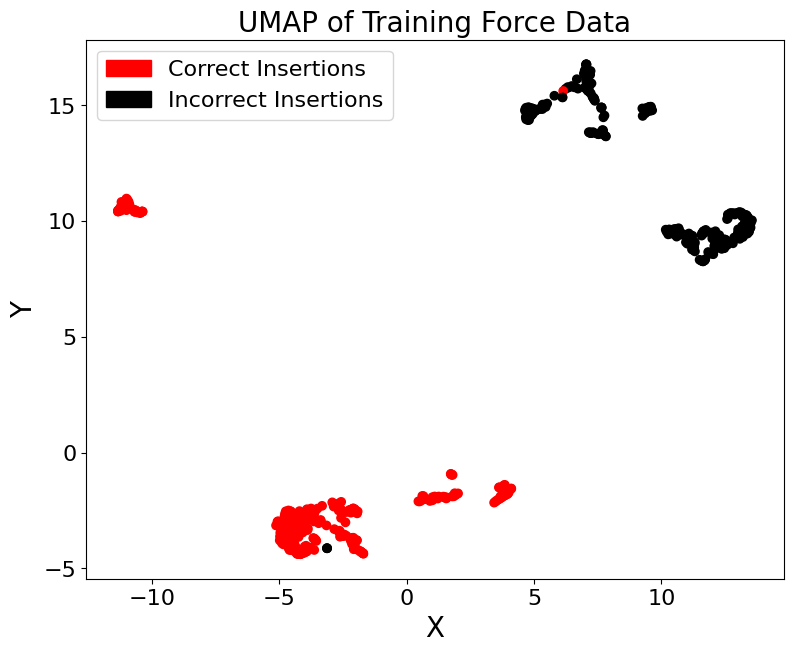}
    \caption{Visualization of UMAP dimensionality reduction on training data from a dimension of 1000 to 2.}
    \label{fig:trainingUmap}
\end{figure}

Firstly, UMAP constructs a high-dimensional neighbor graph that captures local structure by connecting each point to its nearest neighbors according to the chosen distance metric. This graph is then converted into a fuzzy topological representation. Subsequently. UMAP performs a low-dimensional embedding of the data, which involves mapping the data from its original high-dimensional space to a lower-dimensional space, while preserving the neighborhood relationships and global structure.
Fig. \ref{fig:trainingUmap} shows a visualization example of our training data at the end of an experiment.

\begin{figure*}[t]
    \vspace{3mm}
    \centering
    \includegraphics[width=0.95\linewidth]{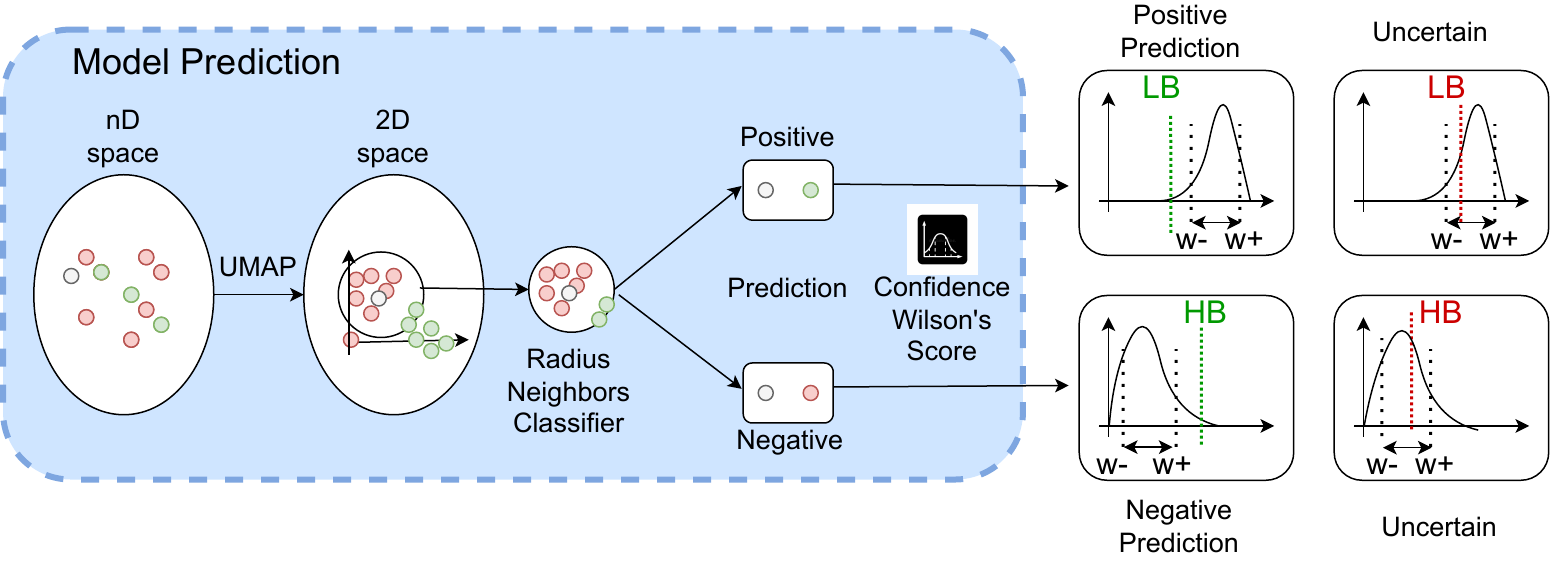}
    \caption{Illustration of the "Model Prediction" component. It combines dimensionality reduction (UMAP), classification (Radius Neighbors), uncertainty quantification (Wilson score) to make predictions more reliable, with "Expensive Verification" fallback when confidence is low.}
    \label{fig:methodology}
\end{figure*}

\subsection{Radius Neighbors Classifier}
Regarding our model prediction selection, we employ the Radius Neighbors Classifier \cite{scikitlearn2011} from Scikit-Learn due to its ease of integration into the engine and its simplicity of interpretation. It is a variant of the $k$-nearest neighbors ($k$-NN) algorithm that, instead of using a fixed number of neighbors $k$, it considers all neighbors within a specified radius $r$ around a query point, assigning the class based on the majority density within that radius. This allows the query point to be compared only with its local neighbors, rather than with distant points that might otherwise be considered nearest neighbors in standard $k$-NN.

\subsection{Confidence Prediction}
\label{subsec:confidenceprediction}
The baseline method for confidence prediction is the Binomial Interval score \cite{lawrence2001}. This is a confidence interval estimator for success probabilities when only the number of trials and observed successes are available. Its formulation is provided in Eq. \ref{eq:BI}, where $p$ denotes the confidence bounds, $n_{s}$ the number of successful trials, $n_{f}$ the number of failed trials, and $k$ is the total number of trials, computed as $n_{s}+n_{f}$. The term $z_{\alpha}$ corresponds to the $1 - \frac{\alpha}{2}$ quantile of the standard normal distribution associated with the target error rate $\alpha$.

\begin{equation}
\label{eq:BI}
    p = \frac{n_s}{k} 
    \pm 
    \frac{z_{\alpha}}{\sqrt{k}} 
    \sqrt{
    \frac{n_s n_f}{k^2}
    } 
\end{equation}

An improvement compared to the Binomial Interval score is the Wilson Score Interval, which performs better with a small sample size or when the success probability is close to 0 or 1. It's formula is different and shown in Eq. \ref{eq:WS}:
\begin{equation}
\label{eq:WS}
    p = \frac{n_s + \frac{1}{2} z_{\alpha}^2}{k + z_{\alpha}^2} 
    \pm 
    \frac{z_{\alpha}}{k + z_{\alpha}^2} 
    \sqrt{
    \frac{n_s n_f}{k} + \frac{z_{\alpha}^2}{4}
    } 
\end{equation}

As shown from computations in Tab. \ref{tab:wilsonformulaincreasedcertainty} and \ref{tab:wilsonformulaincreasedpositive}, when the error tolerance $\alpha$ is reduced, the confidence level increases, which in turn enlarges the gap between the lower and upper bounds of the Wilson interval. In contrast, as the number of identical classifications within the fixed radius increases, both the lower and upper bounds shift upward, while the interval width decreases.


From the user’s perspective, the system guarantees that precision does not fall below the lower bound threshold ($LB$) with a certainty level of $(100 - \alpha)\%$, while the false positive rate remains below the $(100 - LB)\%$ threshold. As an example, with $\alpha=0.05$ and $LB=95\%$ only the last row of Tab. \ref{tab:wilsonformulaincreasedpositive} can satisfy these conditions. This means that, given these conditions, only when approximately all 100 neighbors belong to the same class, the model predicts with high confidence, without the need to rely on the expensive verification.

Depending on the user task requirements and tolerable error, these two parameters can be tuned accordingly.


\begin{table}[h]
    \caption{Wilson Formula results under different conditions. (a) Lower $\alpha$ parameter (b) Increased positive samples in the neighborhood.}
    \centering
    
    \begin{subtable}{\columnwidth}
        \centering
        \begin{tabular}{|c|c|c|c|}
        \hline
        $\alpha$ & $n_s$ & $k$ & Wilson Intervals \\
        \hline
        0.2 & 95 & 100 & [0.91, 0.97]\\ 
        \hline
        0.1 & 95 & 100 & [0.90, 0.97]\\ 
        \hline
        0.05 & 95 & 100 & [0.88, 0.97]\\ 
        \hline
        0.01 & 95 & 100 & [0.86, 0.98]\\ 
        \hline
        \end{tabular}
        \vspace{0.2cm}
        \caption{}
        \label{tab:wilsonformulaincreasedcertainty}
    \end{subtable}
    
    
    \begin{subtable}{\columnwidth}
        \centering
        \begin{tabular}{|c|c|c|c|}
        \hline
        $\alpha$ & $n_s$ & $k$ & Wilson Intervals \\
        \hline
        0.05 & 96 & 100 & [0.901, 0.984]\\ 
        \hline
        0.05 & 97 & 100 & [0.915, 0.989]\\ 
        \hline
        0.05 & 98 & 100 & [0.929, 0.994]\\ 
        \hline
        0.05 & 99 & 100 & [0.945, 0.998]\\ 
        \hline
        0.05 & 100 & 100 & [0.963, 1.000]\\ 
        \hline
        \end{tabular}
        \vspace{0.2cm}
        \caption{}
        \label{tab:wilsonformulaincreasedpositive}
    \end{subtable}
    \label{tab:wilson_combined}
\end{table}

\subsection{Data engine}
After the Wilson score outputs the confidence bounds, we then evaluate the lower bound value. If it exceeds the chosen threshold $LB$, it indicates that the UMAP transformation places the sample in a region densely populated by similar points, allowing the model to predict its class with greater confidence.

In case of a prediction of an incorrect insertion, the object is returned to the bin, and the system proceeds to the next sample. Otherwise, we proceed with the standard sequence of robot actions. If a problem arises during this step (in our case, a system stop), we know that the classification previously made resulted in a false positive, leading to a system stop where the user needs to interfere to restart the process. Consequently, we can save this sample into the database, taking into account that it translates into additional runtime.

After periodic insertion iterations, the system's model can be retrained using the growing database consisting of samples for which the model exhibited low confidence, as well as false positive samples obtained from the "Store in Database" component. 

The data engine has been tested only in CPU, and training time has been measured to be less than a full robotic cycle time, which is on average of $0.45$ seconds with the current dataset size.

\begin{figure*}[thb]
    \vspace{3mm}
    \centering
    \setlength{\fboxsep}{0pt} 
    \setlength{\fboxrule}{4pt} 
    \begin{tabular}{ccc} 
        \includegraphics[width=0.31\textwidth]{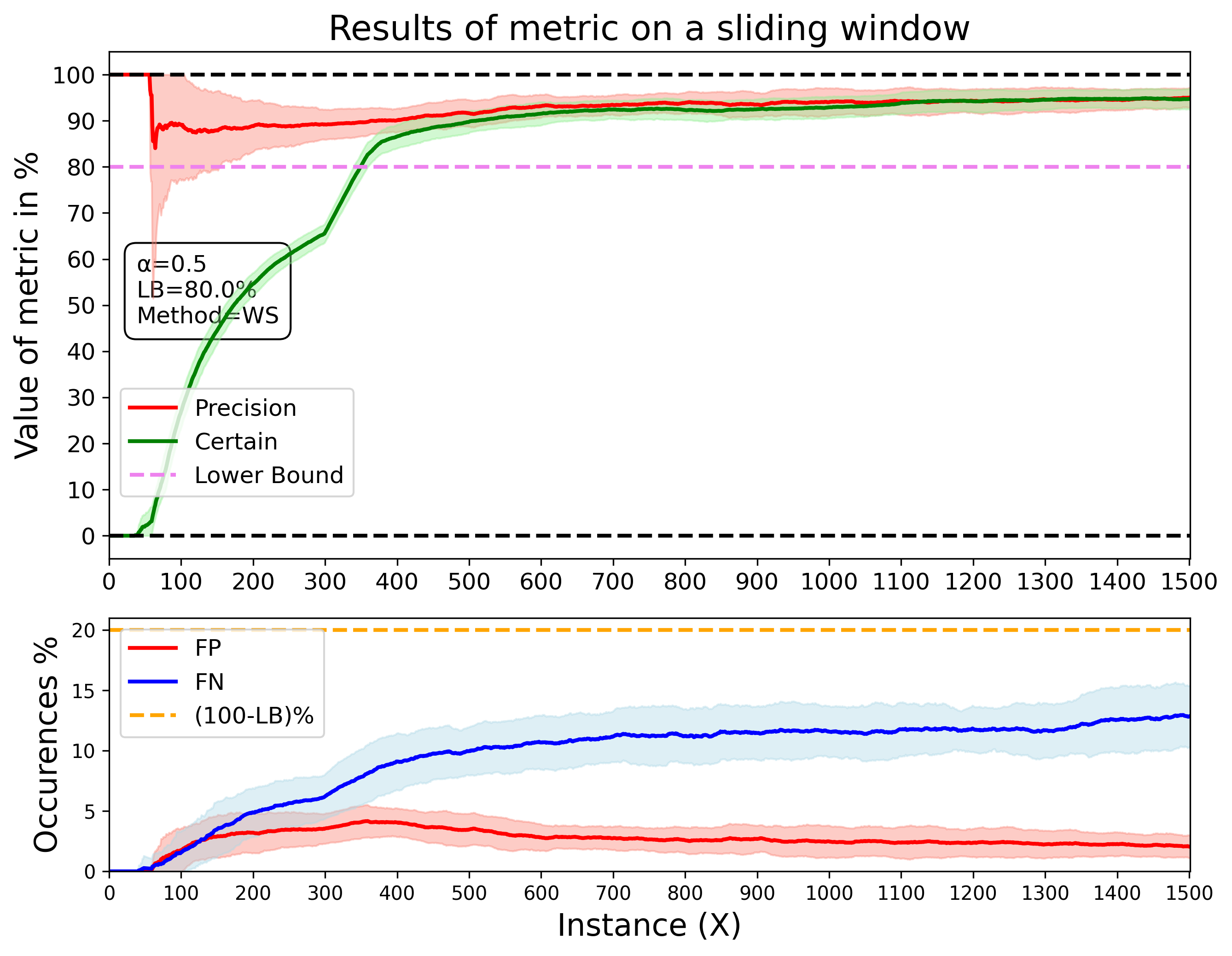} &
        \includegraphics[width=0.31\textwidth]{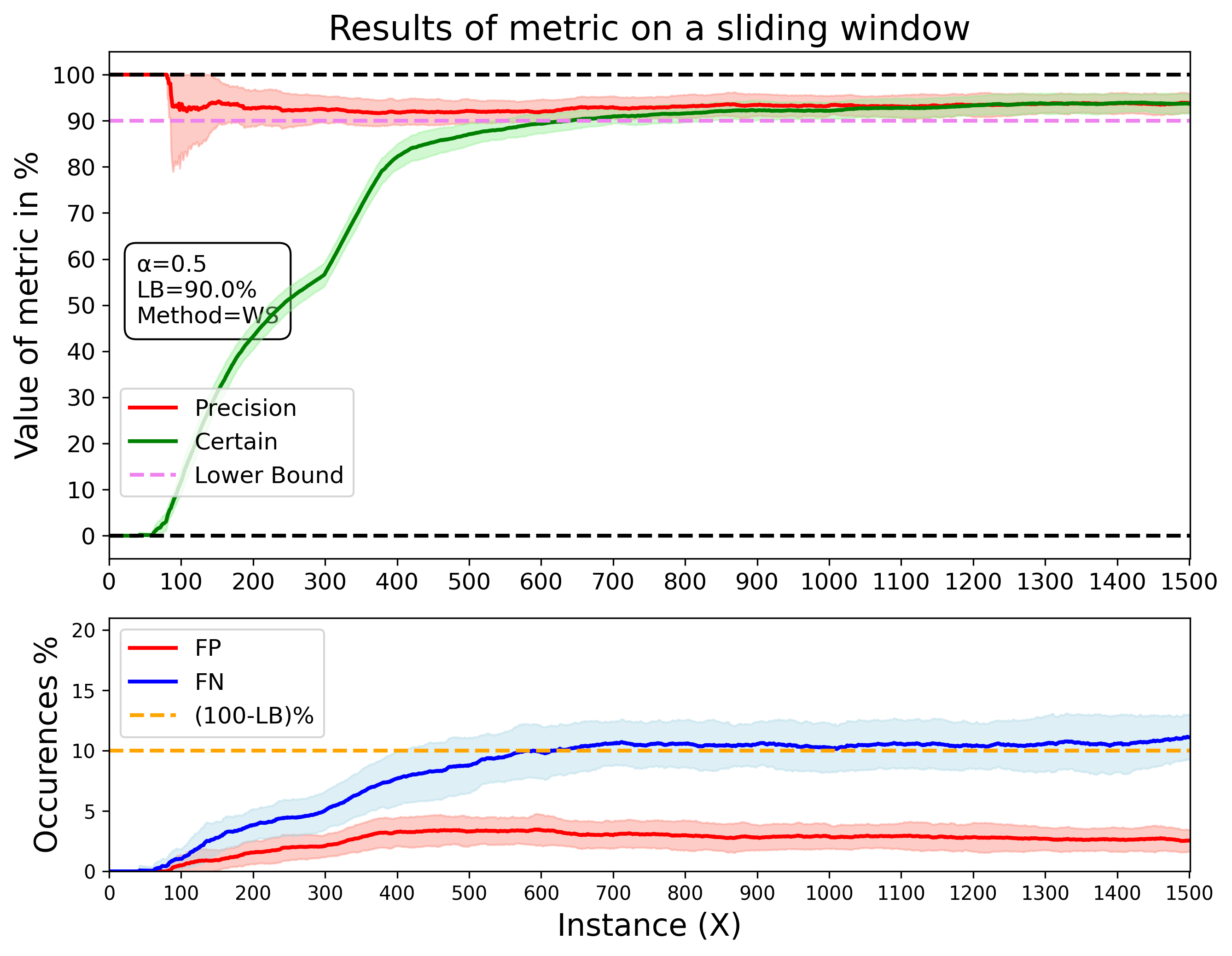} &
        \includegraphics[width=0.31\textwidth]{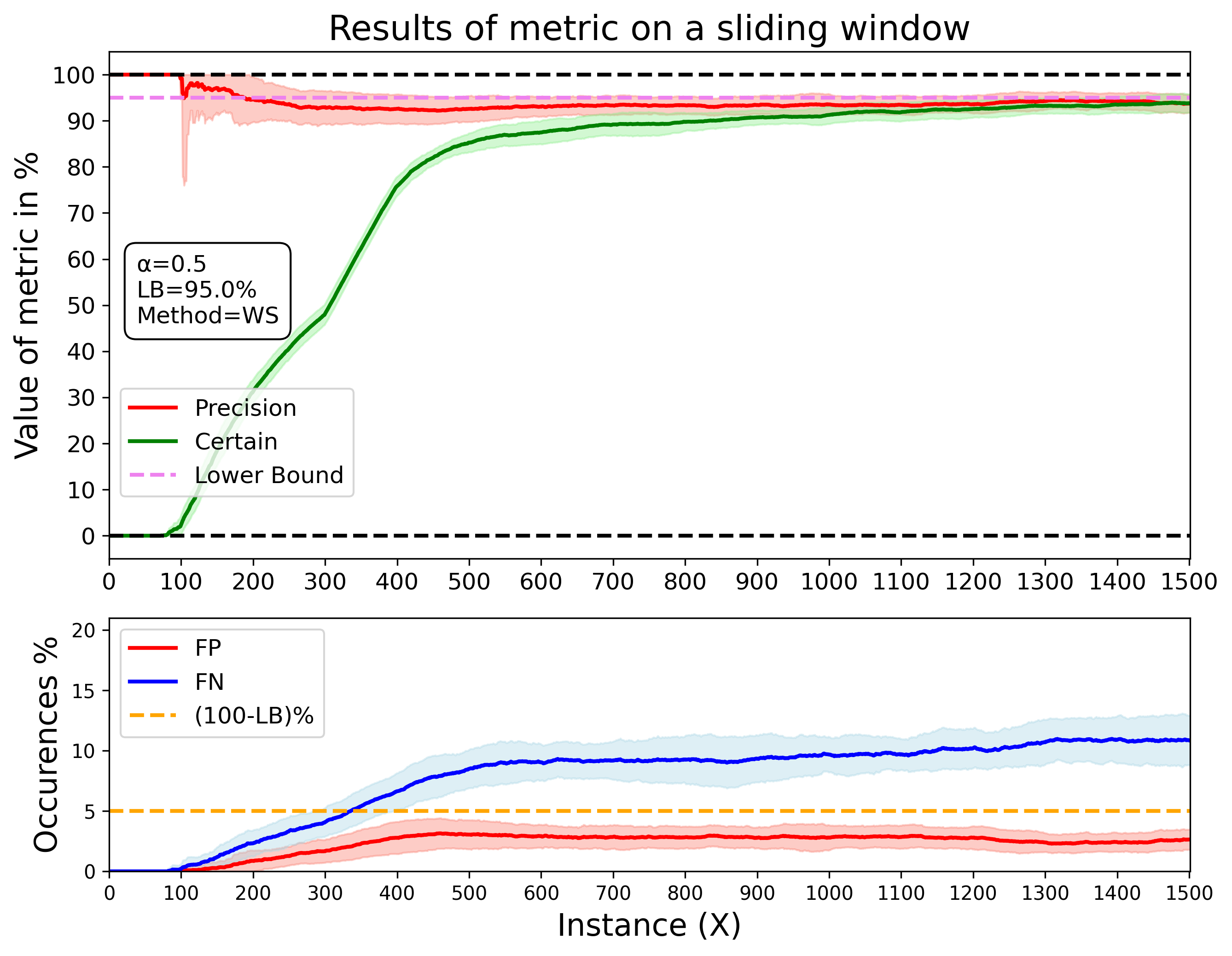} \\
        \includegraphics[width=0.31\textwidth]{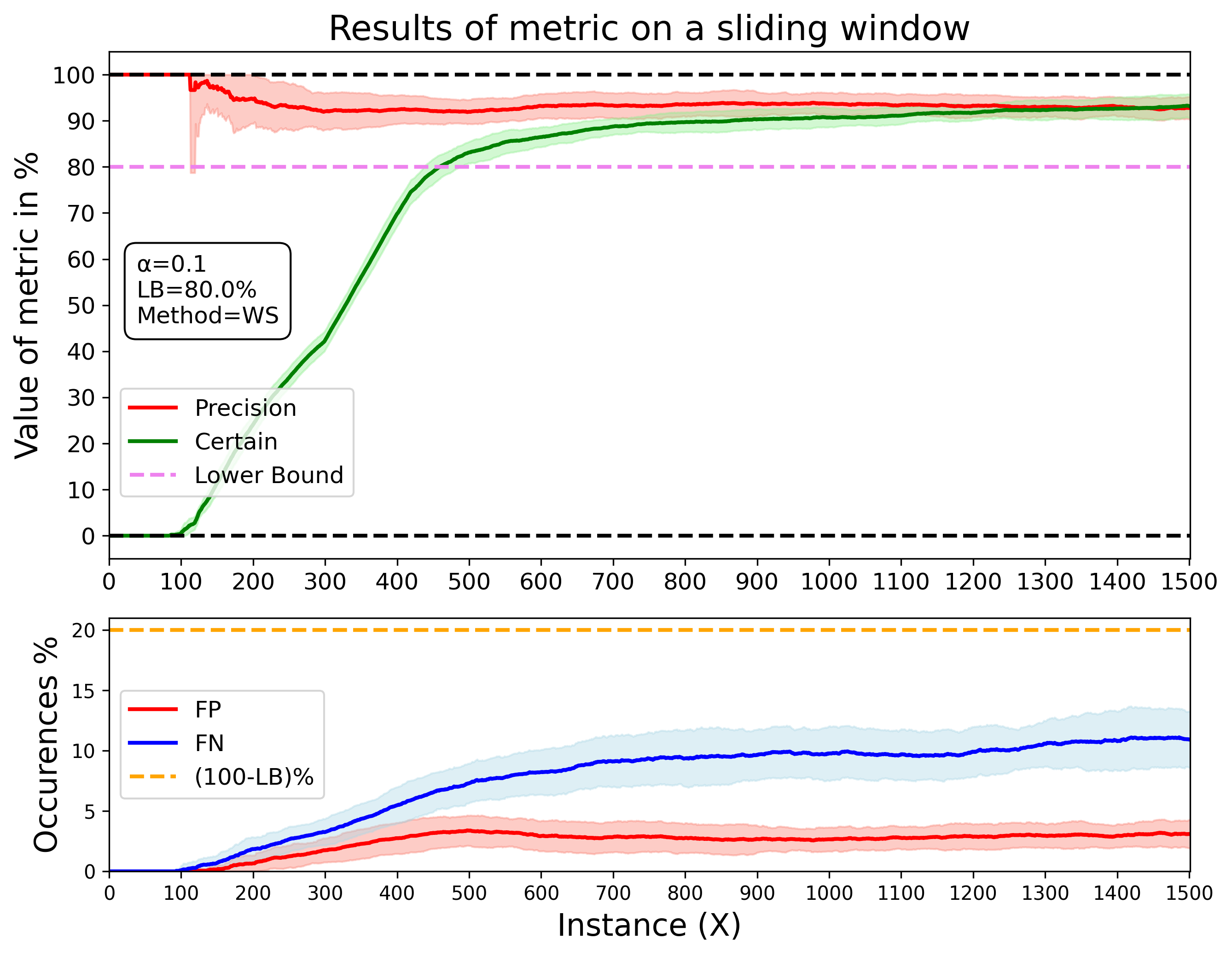} &
        \includegraphics[width=0.31\textwidth]{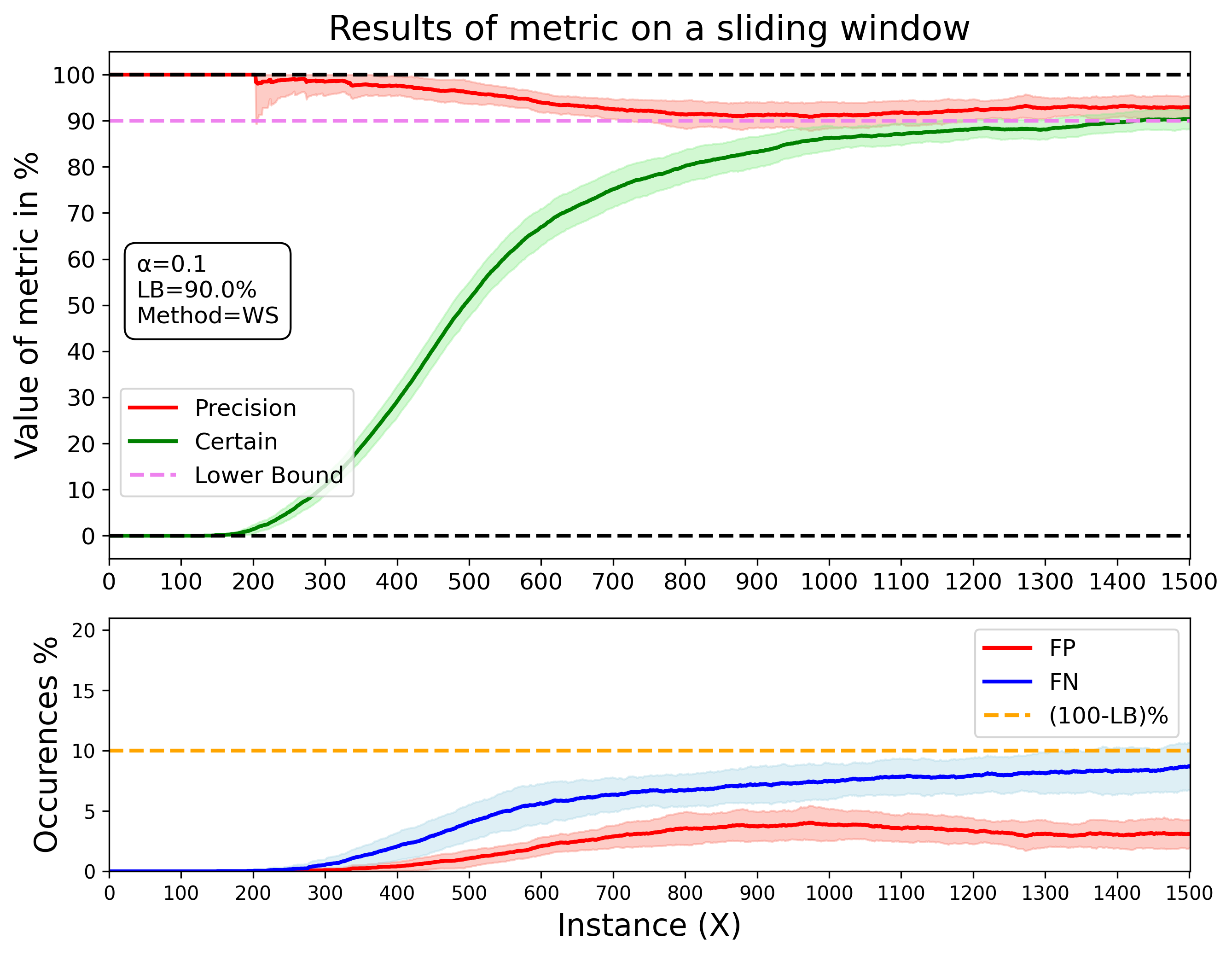} &
        \includegraphics[width=0.31\textwidth]{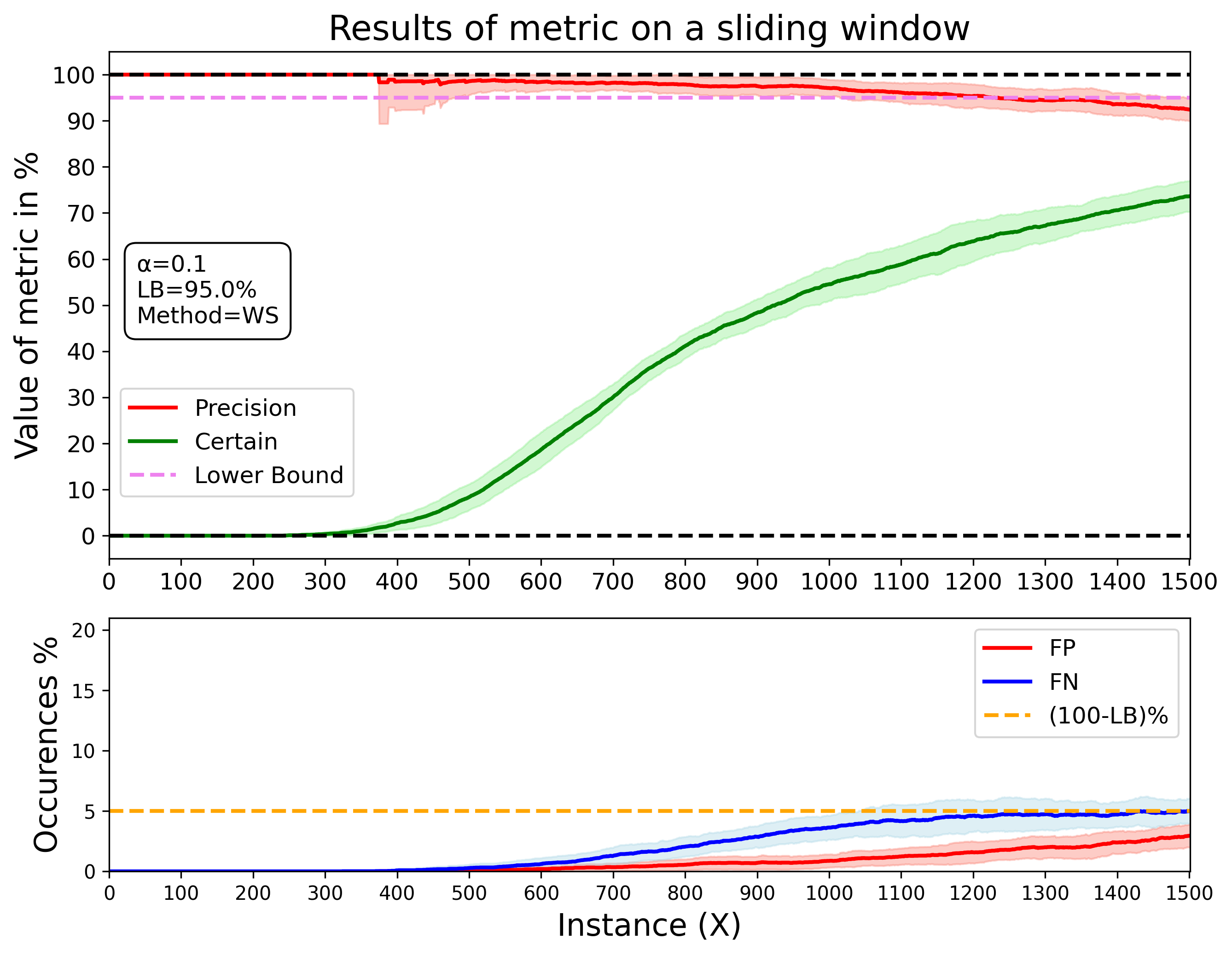} \\
        \includegraphics[width=0.31\textwidth]{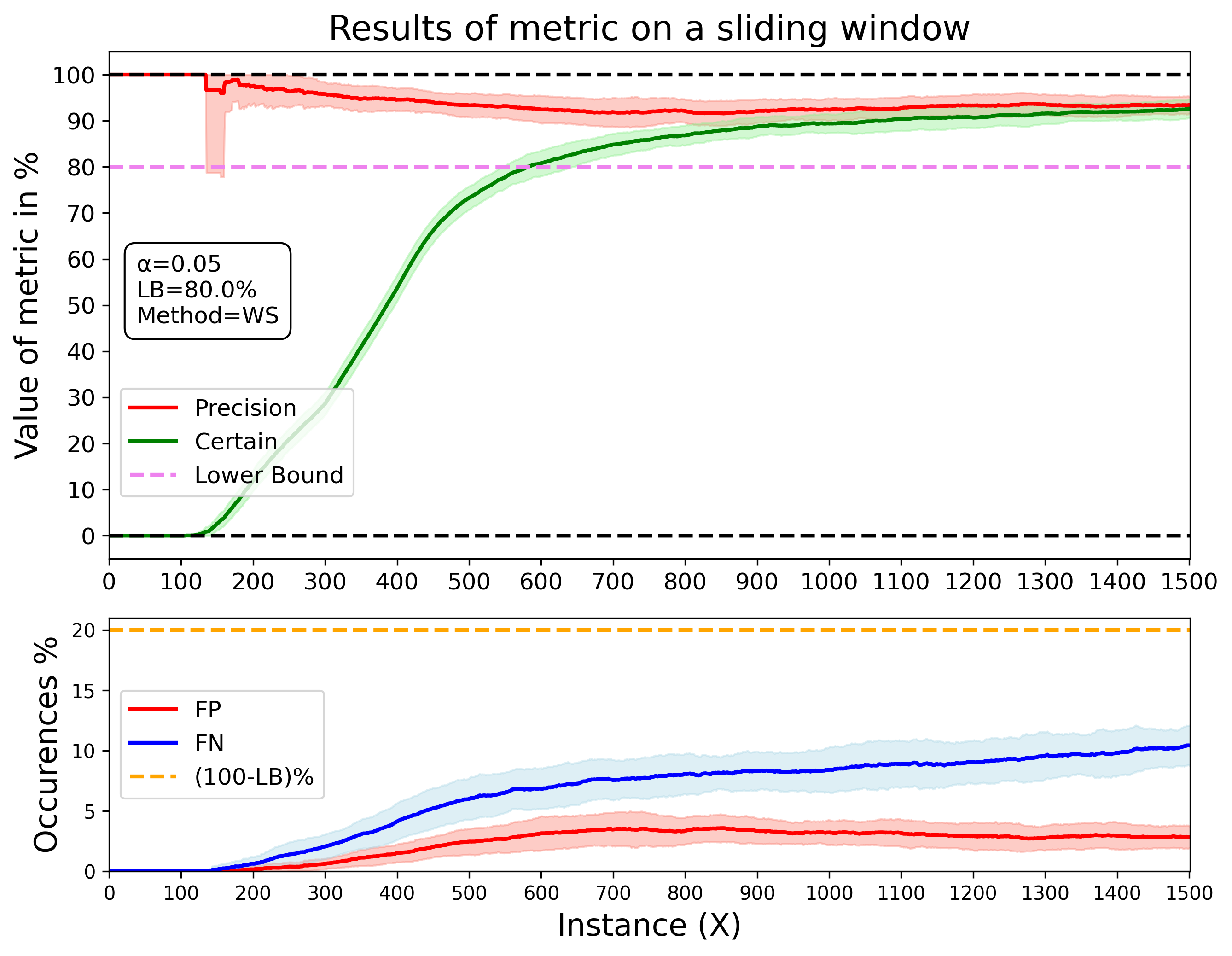} &
        \includegraphics[width=0.31\textwidth]{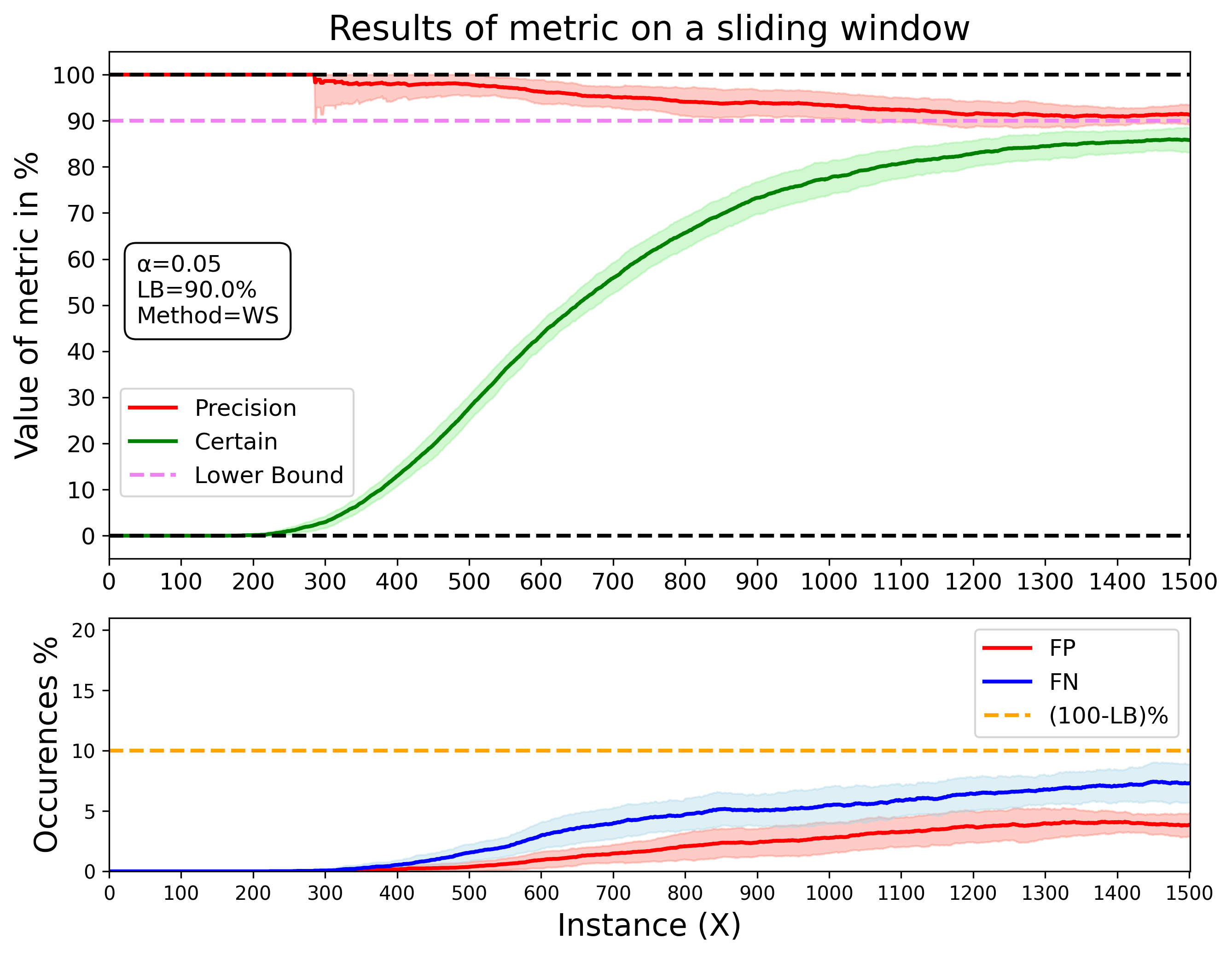} &
        \includegraphics[width=0.31\textwidth]{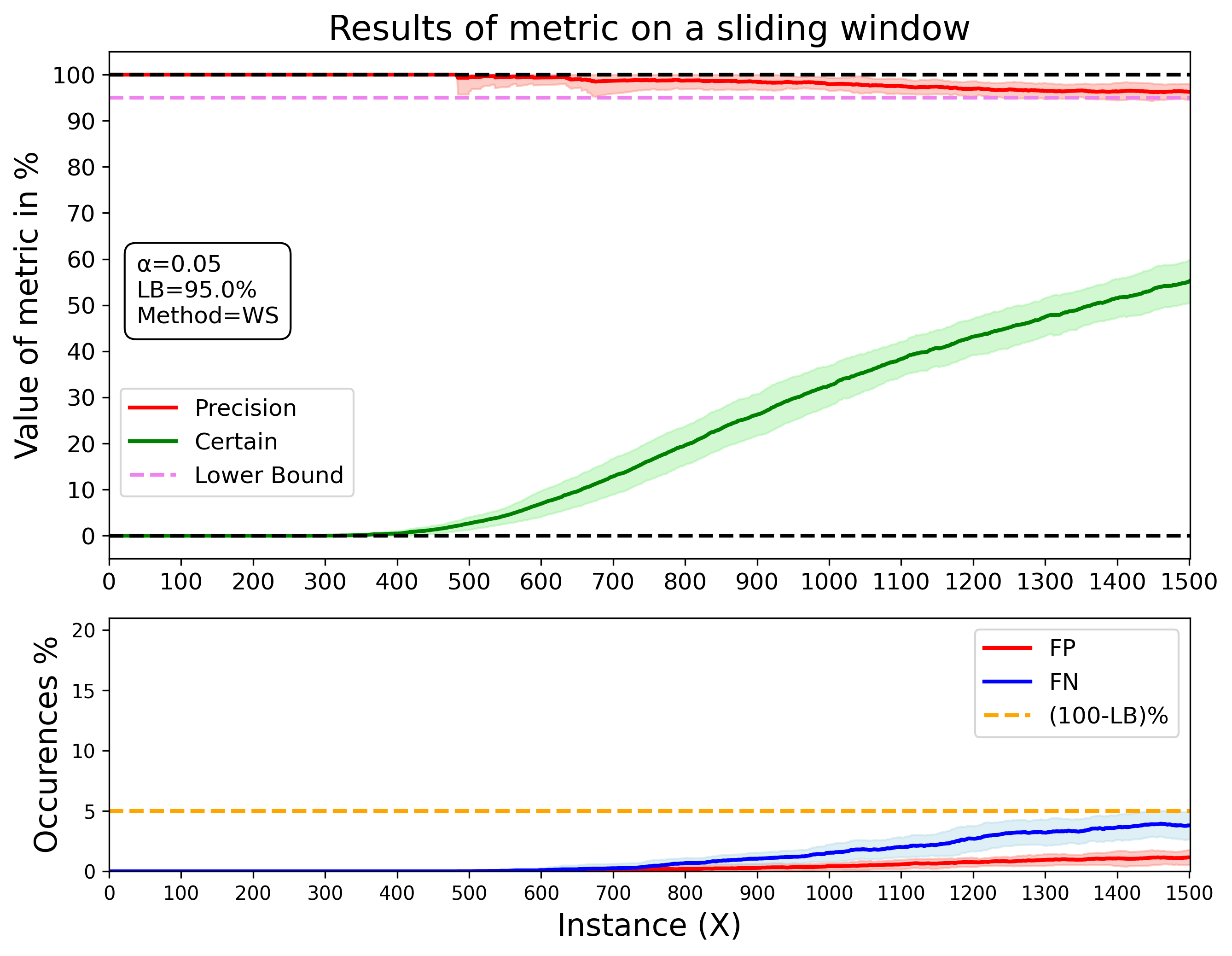} \\
    \end{tabular}
    \caption{Results obtained with varying values of the parameters $\alpha$ and $LB$ threshold described in Sec. \ref{subsec:confidenceprediction} using Wilson Score as our Confidence Bound method.
    In the grid, each column corresponds to a fixed $LB$, while each row corresponds to a fixed $\alpha$. For example, with $\alpha=0.1$ and $LB=90\%$, without the "Model Prediction" component, the system would always perform the "Expensive Verification"; however, with the prediction component, after the 1500th sample, the system would only use the "Expensive Verification" $100-certain\approx10\%$ of the time, while having possible errors in less than $(100-LB)=10\%$ of the time.}

    
    \label{fig:gridWS}
\end{figure*}

\section{Experiments} 
We collected a total of $1503$ force data samples, comprising $803$ positive and $700$ negative insertions. 
Each insertion required between 45 to 50 seconds to complete, meaning that on average it took around 19 hours 49 minutes and 53 seconds in total to collect all the data. It needs to be mentioned that the system could collect data faster, but for debugging reasons and simplicity, the pipeline was not optimized.


We evaluated our approach in a prototype online environment that operates as follows: for each trial, the insertion force dataset is shuffled, and each force data instance is sequentially fed into the workflow shown in Fig. \ref{fig:method}.

For the same configurations of $\alpha$ and $LB$, we repeat the environment 30 times with the shuffled data, and then average the results across these iterations, plotting them over a sliding window of 300 instances.

The results obtained are shown in Fig. \ref{fig:gridWS}, \ref{fig:resBI} and Tab. \ref{tab:resultsWS_BI_combined}. Each figure has two parts: a plot for the precision metric and number of predictions made in percentage over the number of instances of the sliding window at the top, and at the bottom, the percentage of false positive and false negative occurrences over the sliding window.


In the following, we present the outcome obtained using the Wilson Score as the confidence bound method, and an ablation study comparing these results with the baseline performance using the Binomial Interval score.

\begin{table}[h]
    \caption{Mean False Positives count relative to the number of Certain Predictions made across different configurations when using as confidence bound method: a) Wilson Score and b) Binomial Interval score.}
    \centering
    
    \begin{subtable}{\columnwidth}
        \centering
        \begin{tabular}{|c|c|c|c|}
        \hline
        $\alpha$ / Threshold & $80\%$ & $90\%$ & $95\%$ \\
        \hline
        0.5 & 36.1/1317.1 & 39.6/1281.6 & 39.9/1241.4\\ 
        \hline
        0.1 & 40.4/1217.3 & 36.9/1019.8 & 16.9/621.1\\ 
        \hline
        0.05 & 39.6/1151.4 & 30.4/870.2 & 6.5/397.3\\ 
        \hline
        \end{tabular}
        \vspace{0.2cm}
        \caption{Wilson Score}
        \label{tab:wsFalsePositiveCount}
    \end{subtable}
    
    
    \begin{subtable}{\columnwidth}
        \centering
        \begin{tabular}{|c|c|c|c|}
        \hline
        $\alpha$ / Threshold & $80\%$ & $90\%$ & $95\%$ \\
        \hline
        0.5 & 37.8/1322.7 & 43.5/1325.8 & 40.6/1322.5\\ 
        \hline
        0.1 & 40.9/1324.4 & 41.7/1325.6 & 40.5/1325.0\\ 
        \hline
        0.05 & 41.7/1328.2 & 39.7/1324.1 & 40.2/1324.6\\ 
        \hline
        \end{tabular}
        \vspace{0.2cm}
        \caption{Binomial Interval Score}
        \label{tab:biFalsePositiveCount}
    \end{subtable}
    \label{tab:resultsWS_BI_combined}
\end{table}

\subsection{Results: Wilson Score}
The experimental results are presented in Fig. \ref{fig:gridWS} where each plot corresponds to a different configuration of $\alpha$ and $LB$. The mean absolute counts of false positives relative to the number of predictions across different configurations using the Wilson score are summarized in Tab. \ref{tab:wsFalsePositiveCount}, showing that the number of tolerated false positives can be controlled by adjusting $\alpha$ and $LB$.

For a fixed $\alpha$ and different $LB$, precision is consistently above the threshold. 
The $LB$ directly influences the precision lower bound: higher threshold yields higher precision with fewer false positives, but at the cost of a reduced number of overall predictions and by that higher run-times.

Instead, $\alpha$ determines when the model begins to predict by directly influencing the Wilson Score bounds, as shown in Tab. \ref{tab:wilsonformulaincreasedcertainty}. Lower values of $\alpha$ delay predictions by reducing the lower bound, prompting the engine to accumulate more data before issuing confident predictions.

\subsection{Ablation study: Baseline results using Binomial Interval Score}
\begin{figure}
    \vspace{3mm}
    \centering
    \includegraphics[width=0.99\linewidth]{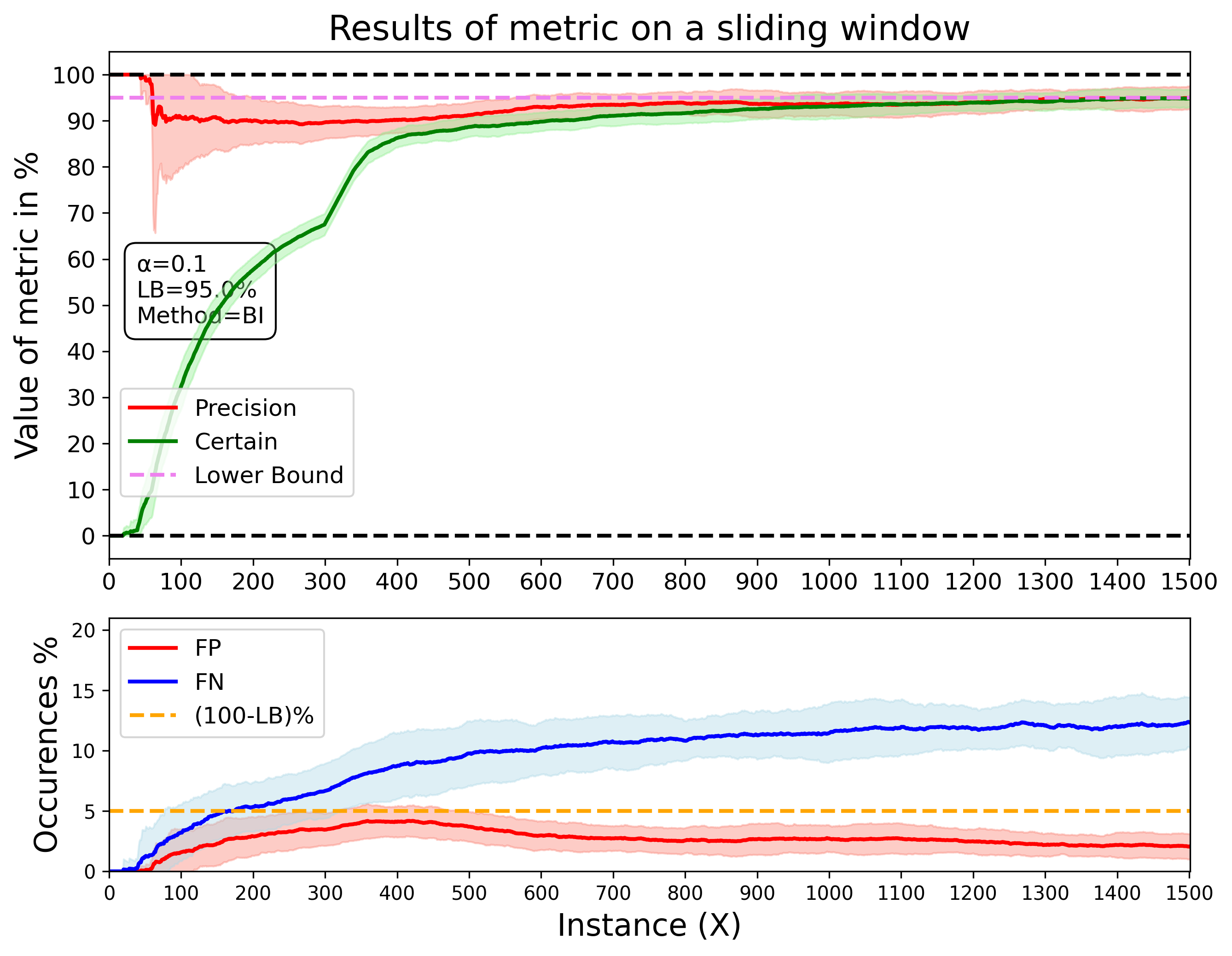}
    \caption{Example of Prototype Online Environment using Binomial Interval score with $\alpha=0.1$ and  $LB=95\%$.}
    \label{fig:resBI}
\end{figure}

We compare our results with the baseline method using Binomial Interval score. An example result can be seen in Fig. \ref{fig:resBI}, and in Tab. \ref{tab:biFalsePositiveCount} the results of different tested configurations are listed.

Varying $\alpha$ and the $LB$ parameter did not yield significant differences. Specifically, using Eq. \ref{eq:BI}, if the new sample has only one neighbor, the lower bound already reaches its maximum value of $p = [1.0, 1.0]$. In contrast, applying Eq. \ref{eq:WS} to this case results in $p = [0.2, 1.0]$, highlighting that the number of neighbors is a critical factor.

When comparing the two methods, the Wilson Score approach tends to reduce reliance on the expensive verification as confidence accumulates. This results in fewer false positives and a higher percentage of certain predictions. 



\section{Conclusion}

This work has proposed a novel data engine for binary classification tasks. The method uses two different verification methods \textit{"Expensive Verification"} and \textit{"Model Prediction"} to obtain both high reliability and fast run-time. By using the Wilson-Score method, confidence bounds on the model prediction can be obtained. By selecting the expensive verification if the confidence is out of the set confidence bounds, the error-rate of the system is controlled.

The method is tested on a real task in a robotic set up. Here we demonstrate that the error rate can be effectively controlled with the $\alpha$ and $LB$ parameters, confidence and lower-bound respectively. It is shown that the average percentage of False Positives is always below the $(100 - LB)\%$ threshold.
Moreover, we demonstrated that after the initial configuration, the data engine can operate autonomously without human intervention for force data collection and labeling. In certain cases, such as with $\alpha = 0.1$ and $LB = 90\%$, the proportion of certain predictions approaches 90\%, indicating that the expensive verification is largely unnecessary in later stages, therefore allowing to reduce the overall system cycle time.

\section{Future Work}

In future research, we plan to extend our approach to tasks with variation in parts, fixture, materials, and other domains that are concerned with similar problems to demonstrate the generalizability of the data engine. As long as the input data for classification is clearly defined and the expensive verification component provides reliable and accurate labeling, the proposed data engine can be systematically applied.


Since machine learning systems deployed across various domains often face persistent challenges, such as overfitting to biased or unrealistic training data and limited ability to generalize to unseen cases, a confidence-aware methodology can help mitigate some of these risks. For example, in medicine, ML models frequently encounter issues related to data quality, model interpretability, and generalization, as medical datasets often contain noise, missing values, and inconsistencies that can significantly impact model performance and reliability \cite{AIMedicine}. By integrating Wilson’s confidence score, the system not only generates predictions but also quantifies their statistical reliability. Consider a medical image classification system trained to detect tumors, the training data may be biased or limited, causing the model to produce uncertain predictions on new images. If the confidence score for a prediction falls below a certain threshold, indicating low statistical reliability, the prediction can be flagged for manual verification by a radiologist, preventing blind trust in uncertain outputs. Furthermore, these flagged cases can be collected and used to retrain the model, reducing overfitting and improving generalization over time. 

\section*{ACKNOWLEDGMENT}
This project was funded in part by Innovation Fund Denmark through the project FERA (3149-00014A), and in part by the SDU I4.0-Lab.

\bibliographystyle{IEEEtran} 
\bibliography{root}


\end{document}